\documentclass[12pt]{article}
\usepackage[margin=1in]{geometry}
\usepackage[T2A]{fontenc}
\usepackage{graphicx} 
\usepackage{amssymb}		
\usepackage{hyperref}	
\usepackage{multirow}
\usepackage{graphicx}
\usepackage{color}
\usepackage{float}
\usepackage{csquotes}
\usepackage{subcaption}
\usepackage{authblk}

\usepackage{multirow}
\usepackage[numbers]{natbib}

\title{Lessons Learned on Information Retrieval in Electronic Health Records: A Comparison of Embedding Models and Pooling Strategies}

\author[1,$\ast$]{Skatje Myers}
\author[2,3]{Timothy A. Miller}
\author[4]{Yanjun Gao}
\author[1]{Matthew M. Churpek}
\author[1]{Anoop Mayampurath}
\author[5]{Dmitriy Dligach}
\author[1]{Majid Afshar}
\affil[1]{University of Wisconsin-Madison}
\affil[2]{Boston Children’s Hospital}
\affil[3]{Harvard Medical School}
\affil[4]{University of Colorado Anschutz}
\affil[5]{Loyola University Chicago}
\date{}

\begin{document}
\maketitle

\abstract{\textbf{Objective:} Applying large language models (LLMs) to the clinical domain is challenging due to the context-heavy nature of processing medical records. Retrieval-augmented generation (RAG) offers a solution by facilitating reasoning over large text sources. However, there are many parameters to optimize in just the retrieval system alone. This paper presents an ablation study exploring how different embedding models and pooling methods affect information retrieval for the clinical domain.\\ \textbf{Methods}: Evaluating on three retrieval tasks on two electronic health record (EHR) data sources, we compared seven models, including medical- and general-domain models, specialized encoder embedding models, and off-the-shelf decoder LLMs. We also examine the choice of embedding pooling strategy for each model, independently on the query and the text to retrieve.\\ \textbf{Results}: We found that the choice of embedding model significantly impacts retrieval performance, with BGE, a comparatively small general-domain model, consistently outperforming all others, including medical-specific models. However, our findings also revealed substantial variability across datasets and query text phrasings. We also determined the best pooling methods for each of these models to guide future design of retrieval systems.\\ \textbf{Discussion}: The choice of embedding model, pooling strategy, and query formulation can significantly impact retrieval performance and the performance of these models on other public benchmarks does not necessarily transfer to new domains. Further studies such as this one are vital for guiding empirically-grounded development of retrieval frameworks, such as in the context of RAG, for the clinical domain.\\  }

\section{Introduction}

Large language models (LLMs) have demonstrated remarkable performance on a wide range of natural language processing tasks, showcasing their potential to advance various domains. However, bringing this benefit to the clinical domain poses significant challenges. The number of progress reports, radiology reports, and other clinical notes in the electronic health record (EHR) that build up over the course of a patient's hospitalization can quickly exceed most current LLM context windows. Furthermore, the utilization of the full context window can cause LLMs to suffer from the \textit{lost-in-the-middle} effect \citep{liu2024lost}, where their ability to utilize information towards the middle of the context decreases as the length of the text increases.

Retrieval-augmented generation (RAG) \citep{lewis2020retrieval} has emerged as a promising technique for enabling reasoning over large text sources. This approach allows for the retrieval of relevant passages to provide as context within the prompt for the generated response. This reduces the prompt size and has also been shown to enhance accuracy in various applications dealing with expansive textual data \citep{gao2023retrieval}.

While the generative LLM component can be easily upgraded as new models are released, creating the vector database that stores the embedded documents is an expensive investment at scale and not as trivially replaceable. It’s therefore vital that the decisions made in designing the retrieval pipeline are well-grounded. For example, one must select a suitable model to create text embeddings, and while public benchmarks such as Massive Text Embedding Benchmark (MTEB) \citep{muennighoff2022mteb} exist, there is no guarantee that the highest performing models were not influenced by data contamination \citep{sainz-etal-2023-nlp} or that they perform well for other domains or texts of other lengths. Furthermore, the optimal choice of embedding pooling method may vary depending on factors such as model architecture, length of text, and the nature of the text.

\begin{figure}[h]
  \centering
  \includegraphics[width=\textwidth]{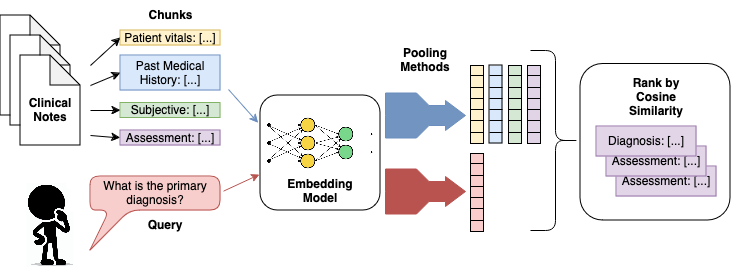}
  \caption{Process of embedding and querying clinical notes.}
  \label{fig:pipeline}
\end{figure}

In this paper, we aim to provide a better understanding of the effects of some of these early decisions on the performance of information retrieval for the clinical domain. This pipeline is illustrated in Figure \ref{fig:pipeline}. We provide an ablation study of recent models and varying embedding pooling methods on three extractive tasks and examine the reproducibility on two data sources: the publicly available MIMIC-III \citep{johnson2016mimic} dataset and a private EHR dataset.

We provide a thorough testing of embedding pooling methods independently on the query and note text, providing statistically verified recommendations on pooling methods to be used for each tested model on the larger text chunks, though we found the choice of pooling strategy for queries to be less significant. 

In our comparison of embedding models, both those explicitly trained for text representation and decoder-only models, we find that the choice of embedding model significantly impacts retrieval performance, with BGE \citep{bge_embedding}, a comparatively small general-domain model, consistently outperforming all others, including medical-specific models and two models which are ranked higher on MTEB. However, our findings also reveal substantial variability across datasets and query text phrasings, highlighting the difficulty in developing a robust retrieval system for novel datasets and tasks.

\section*{Statement of Significance}

\subsection*{Problem}
Applying large language models to electronic health records is challenging due to the vast amount of text per patient, often exceeding LLM context limits. While retrieval-augmented generation shows promise in addressing this, the optimal configurations for the retrieval step—specifically, embedding models and pooling strategies—remains unclear in the clinical domain.

\subsection*{What is already known}
Public benchmarks exist for evaluating embedding models, but their performance on these benchmarks may not translate directly to other domains. Additionally, optimal pooling strategies for embedding models may vary depending on nature and length of the text to represent. The impact of these factors on information retrieval performance for EHR in particular remains understudied. 

\subsection*{What this paper adds}
This study provides a systematic comparison of seven embedding models, including both general-domain and medical-specific models, across three clinical information retrieval tasks using two distinct electronic health record data sources. Our findings offer empirical evidence on the performance of different embedding models and pooling strategies in the clinical domain, providing guidance for optimizing retrieval systems for EHR and highlighting the need for domain-specific evaluation.

\section{Related Work}
\label{related}

In a similar vein to our work, Aperdannier et al. \cite{10.1007/978-3-031-53963-3_36} provided a rich comparison of embedding models for the search of German-language insurance text. They tested different document splitting methods, chunk sizes, and models. The pooling method was not a variable they included, instead using mean pooling for all experiments. Our experiments tested different models and tasks than their work, which found the closed source OpenAI text-embedding-ada-002 model \citep{openai-embeddings} to perform best. Although their systematic comparison provided valuable insights for German-language insurance text retrieval, the transferability of these findings to clinical contexts remains unclear.

The dearth of best practices for the various components of RAG systems, specifically in the clinical domain, has been recently addressed by the MedRAG toolkit \citep{xiong2024benchmarking}. This toolkit allows for convenient swapping of components -- the text to search over, the retrieval method, and LLMs for generation. They evaluated a number of permutations on their newly proposed Mirage benchmark, which is comprised of five medical question-answering corpora, although none incorporate EHR documents. In contrast, our work is concerned primarily with EHR documents, and our methodological focus is on optimizing the retrieval step before the introduction of the numerous decisions that go into the generative process of the framework.

\section{Methods}
\label{methods}

\subsection{Tasks}
\label{tasks}

We designed three information retrieval tasks to test on two EHR data sources, motivated by future use cases of generating a discharge summary for a hospital encounter or question answering. To evaluate the efficacy of retrieval approaches for these tasks, we developed a semi-automatic approach for generating labeled data.

Each hospital encounter consists of a discharge summary and the unstructured notes for the hospitalization that temporally preceded it. We identified three types of information of interest:

\begin{enumerate}
    \item Primary diagnosis (e.g. \textit{aspiration pneumonia, type 2 diabetes})
    \item Antibiotics (e.g. \textit{amoxicillin, doxycycline})
    \item Invasive/surgical procedures (e.g. \textit{left ICA endarterectomy, flexible bronchoscopy})
\end{enumerate}

For the antibiotic task, we automatically mapped the text to medical concepts from the National Library of Medicine's Unified Medical Langauge System (UMLS) with semantic type T195 (Antibiotics) within the notes using the tool QuickUMLS \citep{soldaini2016quickumls} and treated all such mentions as the target for retrieval.

For the primary diagnosis and surgical procedure tasks, we aimed to simulate the use case of generating a discharge summary through a RAG framework by specifically targeting the ground truth found in the summary. For each encounter, we attempted to extract the primary diagnosis and the surgical procedures sections from the summary using regular expressions. When this information was available, we then identified mentions of the target diagnosis and procedures within the rest of the notes as our retrieval goals.

Due to the frequent use of acronyms and the numerous ways of expressing the same medical concept, we needed to employ a fuzzy matching technique to find these mentions. We first employed QuickUMLS to identify UMLS concepts within the text as potential matches, restricting by appropriate semantic types (Table \ref{tab:sem_types}). In the case of the primary diagnosis, we calculated the cosine similarity between the BioLORD-2023 \citep{10.1093/jamia/ocae029} embedding of the known diagnosis and that of each of the UMLS entity spans. If the similarity was >= 0.6, this was considered a positive match. This method enabled us to correctly identify occurrences such as \textit{``left knee OA''} as a mention of the known primary diagnosis of \textit{``osteoarthritis of the left knee''}.

\begin{table}[H]
\centering
\begin{tabular}{l|l|}
\cline{2-2}
                                           & \textbf{Valid UMLS types}                                                           \\ \hline
\multicolumn{1}{|l|}{\textbf{Diagnosis}}   & \begin{tabular}[c]{@{}l@{}}T047, T046, T191, \\ T190, T184, T033, T037\end{tabular} \\ \hline
\multicolumn{1}{|l|}{\textbf{Antibiotics}} & T195                                                                                \\ \hline
\multicolumn{1}{|l|}{\textbf{Procedures}}  & T061, T060                                                                          \\ \hline
\end{tabular}
\caption{Allowable semantic types when identifying mentions of the target.}
\label{tab:sem_types}
\end{table}

In the case of surgical procedures, where the ground truth section of the discharge summary typically contained more free text, we identified the procedures from the section using QuickUMLS and considered mentions within the rest of the encounter notes to be matches if their BioLORD-2023 embeddings were similar to any of the procedure entities.

We note that it should not be expected for any retrieval method to achieve a perfect score for the ``diagnosis'' and ``procedures'' tasks. The understanding, for instance, of \textit{which} diagnosis is the primary diagnosis is not necessarily represented in text embeddings, merely that the text contains \textit{a} diagnosis. Additionally, not all invasive procedures are noted in the discharge summary and therefore ``incorrect'' procedures mentioned in the text may be ranked highly. In a RAG framework, the generative step would provide this reasoning. The design of these tasks is intended to facilitate using the same datasets for future work that explores the relation between performance on retrieval and the final performance of a RAG system.

\subsection{Datasets}
\label{datasets}

We used two data sources to construct the testing data for each task independently -- private EHR sourced from the University of Wisconsin (UW) hospital and the publicly-available MIMIC-III dataset \citep{johnson2016mimic}.

Our task datasets consist of varying numbers of patient encounters, which are comprised of all available notes prior to the discharge summary for a given hospital encounter. These notes were segmented into chunks of a maximum of 256 token lengths, with a sliding window of 50. To determine the necessary sample size of our datasets, we used the Sample Size Calculator for Evaluations (SLiCE) \citep{canales2021assessing}, which uses predefined confidence intervals and levels to calculate the minimum sample size required for robust metrics of performance that are adequately powered to detect a statistical difference.  With a maximally conservative setting of precision and recall of 0.5 and the variance around the 95\% confidence level set to 0.05, for all six datasets we exceeded the required sample size to meet these criteria by at least 38\%. 

For computational practicality, we limited our consideration of the UW dataset to encounters of five days or less in length of stay. Even with this restriction, the encounters included were comprised of 5,245 to 63,376 tokens each, highlighting the importance of retrieval solutions for the clinical domain. We described the dataset statistics further in Table \ref{tab:data_stats}. There was some variance between the UW data and MIMIC-III in the prevalence of relevant note chunks that contain the target information. Additionally, MIMIC-III typically consisted of fewer tokens than the UW data.

\begin{table}[H]
\centering
\begin{tabular}{l|lll|}
\cline{2-4}
\multicolumn{1}{c|}{\textbf{}}                 & \multicolumn{1}{c|}{\textbf{Diagnosis}} & \multicolumn{1}{c|}{\textbf{Procedures}} & \multicolumn{1}{c|}{\textbf{Antibiotics}} \\ \cline{2-4} 
\multicolumn{1}{c|}{\textbf{}}                 & \multicolumn{3}{c|}{\textbf{MIMIC-III}}                                                                                        \\ \hline
\multicolumn{1}{|l|}{\textbf{\# encounters}}   & \multicolumn{1}{l|}{20}                 & \multicolumn{1}{l|}{15}                  & 15                                        \\ \hline
\multicolumn{1}{|l|}{\textbf{Avg notes/enc.}}  & \multicolumn{1}{l|}{19.7}               & \multicolumn{1}{l|}{25.7}                & 33.4                                      \\ \hline
\multicolumn{1}{|l|}{\textbf{Avg tokens/enc.}} & \multicolumn{1}{l|}{11,569}             & \multicolumn{1}{l|}{15,250}              & 20,012                                    \\ \hline
\multicolumn{1}{|l|}{\textbf{\# chunks}}       & \multicolumn{1}{l|}{3503}               & \multicolumn{1}{l|}{3501}                & 4557                                      \\ \hline
\multicolumn{1}{|l|}{\textbf{Relevant chunks}} & \multicolumn{1}{l|}{18.2\%}             & \multicolumn{1}{l|}{36.1\%}              & 14.9\%                                    \\ \hline
\multicolumn{1}{c|}{\textbf{}}                 & \multicolumn{3}{c|}{\textbf{UW}}                                                                                          \\ \hline
\multicolumn{1}{|l|}{\textbf{\# encounters}}   & \multicolumn{1}{l|}{10}                 & \multicolumn{1}{l|}{10}                  & 10                                        \\ \hline
\multicolumn{1}{|l|}{\textbf{Avg notes/enc.}}  & \multicolumn{1}{l|}{42.7}               & \multicolumn{1}{l|}{46}                  & 47.7                                      \\ \hline
\multicolumn{1}{|l|}{\textbf{Avg tokens/enc.}} & \multicolumn{1}{l|}{24,684}             & \multicolumn{1}{l|}{31,793}              & 29,468                                    \\ \hline
\multicolumn{1}{|l|}{\textbf{\# chunks}}       & \multicolumn{1}{l|}{3,956}              & \multicolumn{1}{l|}{5,208}               & 4,741                                     \\ \hline
\multicolumn{1}{|l|}{\textbf{Relevant chunks}} & \multicolumn{1}{l|}{17.62\%}            & \multicolumn{1}{l|}{10.62\%}             & 11.79\%                                   \\ \hline
\end{tabular}
\caption{Statistics about the six datasets. "Relevant chunks" are those that contain at least one occurrence of the target information, such as the primary diagnosis.}
\label{tab:data_stats}
\end{table}

\subsection{Models and Pooling Methods}

Although there is a wide array of language models available today, practical constraints limit the number of models we were able to evaluate. We aimed to cover a diverse set of models in our study, including both medical- and general-domain models, as well as encoder models specialized for text embeddings and decoder-only architectures.

We included four models designed for embedding representations:
\begin{itemize}
    \item \textbf{BGE-en-large-v1.5} \citep{bge_embedding} (335M parameters): A general-purpose BERT-based embedding model trained through contrastive learning.
    \item \textbf{Gatortron-large} \citep{yang2022large} (8.9B parameters): A clinical BERT model trained on a large amount of EHR and PubMed. Note: A small portion of the pre-training data was text from MIMIC-III. 
    \item \textbf{SFR-Embedding-Mistral} \citep{SFRAIResearch2024}: A further fine-tuned version of E5-Mistral-7B-Instruct \citep{wang2023improving}, which is a fine-tuned Mistral-7B-Instruct trained on synthetic data through contrastive loss.
    \item \textbf{LLM2Vec-Meta-Llama-3-8B-Instruct-mntp-supervised} \citep{behnamghader2024llm2vec}: This model modified the Llama-3-8B-Instruct model to enable bi-directional attention and trained it with their novel masked next token prediction method.
\end{itemize}

and three generative decoder-only models:

\begin{itemize}
    \item \textbf{Llama-3-8B-Instruct} \citep{llama3modelcard}
    \item \textbf{Mistral-7B-Instruct} \citep{jiang2023mistral}
    \item \textbf{BioMistral} \citep{labrak2024biomistral}: A version of Mistral-7B-Instruct which has been further pre-trained on PubMed Central.
\end{itemize}

Due to the datasets containing PHI and being subject to a data use agreement, we did not evaluate on any closed source models.

For each model, we used between 4 and 7 different phrasings of the query per task (see Tables \ref{tab:diag_query}, \ref{tab:proc_query}, and \ref{tab:anti_query} in the Appendix), constructed to be simple and intuitive and without system prompting or extensive tuning in order to provide a generalizable statistical approximation of using these configurations in new use cases.

In order to extract text embeddings from these models, we must pool the last hidden layer. Some models have recommended ways of extracting embeddings. SFR-Embedding-Mistral and LLM2Vec-Llama-3 were both trained to use their own particular query formats, with the embedding derived from either the final token or from mean pooling, respectively. BGE and Gatortron were trained to use a CLS token, with BGE also trained to use a particular query prompt. We go beyond these to consider additional query formats and pooling strategies (\textbf{mean pooling}, \textbf{weighted mean pooling}, \textbf{max pooling}) and assessed them on note chunks and queries independently. For BGE and Gatortron, we tested using the \textbf{CLS token}, but for the rest of the models, we swapped this method with using the \textbf{last token}.

\subsection{Evaluation Plan}

The final step was ranking the note embeddings by cosine similarity to the query embedding and evaluate the ranking by average precision, where a note chunk that contains a mention of the target information is considered a positive instance.

We calculated the success of the various configuration permutations using mean average precision (MAP). The average precision of a ranked list of chunks is an approximation of the Area Under the Precision-Recall Curve. By performing a repeated measures analysis of variance (ANOVA) for each model we found that the pooling method for the note chunks has a significant effect on performance, and therefore aim to control for this in our later comparisons.

For each model, we determined the most robust note pooling strategies across all experiments by performing a post-hoc pairwise Tukey's test between the different strategies to examine the significance of the differences between them.

Through these permutations of models, datasets, queries, query pooling methods, and note pooling methods, we have tested 3,488 configurations on their ability to retrieve the chunks of clinical notes that contain information relevant to the task target.

\section{Results}
\label{results}

In Table \ref{tab:best_pooling}, we present our findings on the best \textit{note} embedding pooling strategies for each model, across all queries and tasks. These results were largely consistent between the datasets. Through the same testing method, we found that the query pooling method has an insignificant effect on performance for most models.

\begin{table}[H]
\centering
\begin{tabular}{l|l|l|}
\cline{2-3}
                                                          & \textbf{Model}        & \textbf{Note Pooling}           \\ \hline
\multicolumn{1}{|l|}{\multirow{7}{*}{\textbf{U of Wisconsin}}}   & BGE-large-en          & weighted mean, CLS, mean        \\ \cline{2-3} 
\multicolumn{1}{|l|}{}                                    & BioMistral            & mean, weighted mean             \\ \cline{2-3} 
\multicolumn{1}{|l|}{}                                    & Gatortron-large       & not statistically significant   \\ \cline{2-3} 
\multicolumn{1}{|l|}{}                                    & LLM2Vec-Llama-3-8B    & not statistically significant   \\ \cline{2-3} 
\multicolumn{1}{|l|}{}                                    & Llama3-8b-Instruct    & weighted mean, mean             \\ \cline{2-3} 
\multicolumn{1}{|l|}{}                                    & Mistral-7B-Instruct   & mean, max, weighted mean        \\ \cline{2-3} 
\multicolumn{1}{|l|}{}                                    & SFR-Embedding-Mistral & weighted mean, last token, mean \\ \hline
\multicolumn{1}{|l|}{\multirow{7}{*}{\textbf{MIMIC-III}}} & BGE-large-en          & weighted mean, mean             \\ \cline{2-3} 
\multicolumn{1}{|l|}{}                                    & BioMistral            & not statistically significant   \\ \cline{2-3} 
\multicolumn{1}{|l|}{}                                    & Gatortron-large       & not statistically significant   \\ \cline{2-3} 
\multicolumn{1}{|l|}{}                                    & LLM2Vec-Llama-3    & mean, weighted mean             \\ \cline{2-3} 
\multicolumn{1}{|l|}{}                                    & Llama3-8b-Instruct    & weighted mean, mean             \\ \cline{2-3} 
\multicolumn{1}{|l|}{}                                    & Mistral-7B-Instruct   & mean, weighted mean             \\ \cline{2-3} 
\multicolumn{1}{|l|}{}                                    & SFR-Embedding-Mistral & last token, weighted mean       \\ \hline
\end{tabular}

\caption{Each model's best embedding pooling methods for the 256-token-maximum note chunks. If multiple methods are listed, we did not find a significant (\textit{p}\textless0.05) difference between them.}
\label{tab:best_pooling}
\end{table}

In Table \ref{tab:results}, we present the mean average precision for the different models, using only the best note pooling strategy for each model, across the various query/query pooling configurations. It should be noted that due to the prevalence of the target information being different between datasets, these scores should not be directly interpreted as whether models perform better on one type of data than the other. With MIMIC-III having a higher prevalence of relevant information for all three tasks, higher scores on that dataset are expected.

\begin{table}[H]
\centering
\begin{tabular}{l|l|l|}
\cline{2-3}
                                                         & U of Wisconsin                    & MIMIC-III                         \\ \hline
\multicolumn{1}{|l|}{BGE-large-en}                       & \textbf{0.403 {[}0.385, 0.421{]}} & \textbf{0.475 {[}0.457, 0.493{]}} \\ \hline
\multicolumn{1}{|l|}{BioMistral}                         & 0.276 {[}0.255, 0.298{]}          & 0.328 {[}0.300, 0.357{]}          \\ \hline
\multicolumn{1}{|l|}{Gatortron-large}                    & 0.191 {[}0.184, 0.198{]}          & 0.270 {[}0.241, 0.298{]}          \\ \hline
\multicolumn{1}{|l|}{Llama3-8B-Instruct}                 & 0.313 {[}0.292, 0.334{]}          & 0.359 {[}0.332, 0.385{]}          \\ \hline
\multicolumn{1}{|l|}{LLM2Vec-Llama-3-8B} & 0.229 {[}0.207, 0.251{]}          & 0.422 {[}0.408, 0.437{]}          \\ \hline
\multicolumn{1}{|l|}{Mistral-7B-Instruct}                & 0.258 {[}0.239, 0.278{]}          & 0.362 {[}0.336, 0.387{]}          \\ \hline
\multicolumn{1}{|l|}{SFR-Embedding-Mistral}              & 0.302 {[}0.283, 0.322{]}          & 0.417 {[}0.394, 0.439{]}          \\ \hline
\end{tabular}
\caption{Mean average precision [95\% CI] for the models across the queries.}
\label{tab:results}
\end{table}

BGE had a mean average precision of 0.403 across the tasks for the UW dataset and 0.475 for the MIMIC datasets. These results were significantly (\textit{p}\textless0.05) better than all other models tested. On the other end of the spectrum, we found Gatortron to perform significantly (\textit{p}\textless0.05) worse than all other models on the UW dataset and all models other than LLM2Vec-Llama-3-8B on the UW dataset. 

The performance differences among the remaining models were less pronounced, with overlapping confidence intervals in some cases. LLM2Vec-Meta-Llama-3-8B-Instruct-mntp-supervised, which was based on Llama3-8B-Instruct, performed significantly better than its based model on the MIMIC-III data, but significantly worse on the UW data. This was further notable because off-the-shelf decoder-only models are not a conventional choice for embedding representations, though Llama-3-8B performs competitively in our evaluations. However, SFR-Embedding-Mistral significantly outperformed Mistral-7B-Instruct, demonstrating the potential benefits of specialized embedding models.

Interestingly, we found no significant difference between the performance of Mistral-7B-Instruct and its medical domain counterpart, BioMistral (\textit{p}=0.91 and 0.50 on UW and MIMIC datasets, respectively).

Overall, most models improved over the baseline of randomly ranking the note chunks, with a few exceptions. After performing one-sample t-tests against the random baseline for each dataset and task, with Bonferroni correction applied to control for multiple comparisons, we found that Gatortron performed significantly \textit{worse} than baseline (adjusted \textit{p}=4.89E-20) for the MIMIC-III ``diagnosis'' task and not significantly different from baseline on UW ``procedures'' (adjusted \textit{p}=1). BioMistral, Mistral-Instruct, and LLM2Vec-Llama-3 also did not differ from the baseline for MIMIC-III ``diagnosis'' (adjusted \textit{p}=1 in all cases), and furthermore, LLM2Vec-Llama-3 performed significantly worse (adjusted \textit{p}=3.45E-18) on UW ``diagnosis'' and did not differ from the baseline (adjusted \textit{p}=1) on UW ``procedures''.

\begin{figure}[H]
  \centering
  \includegraphics[width=\textwidth]{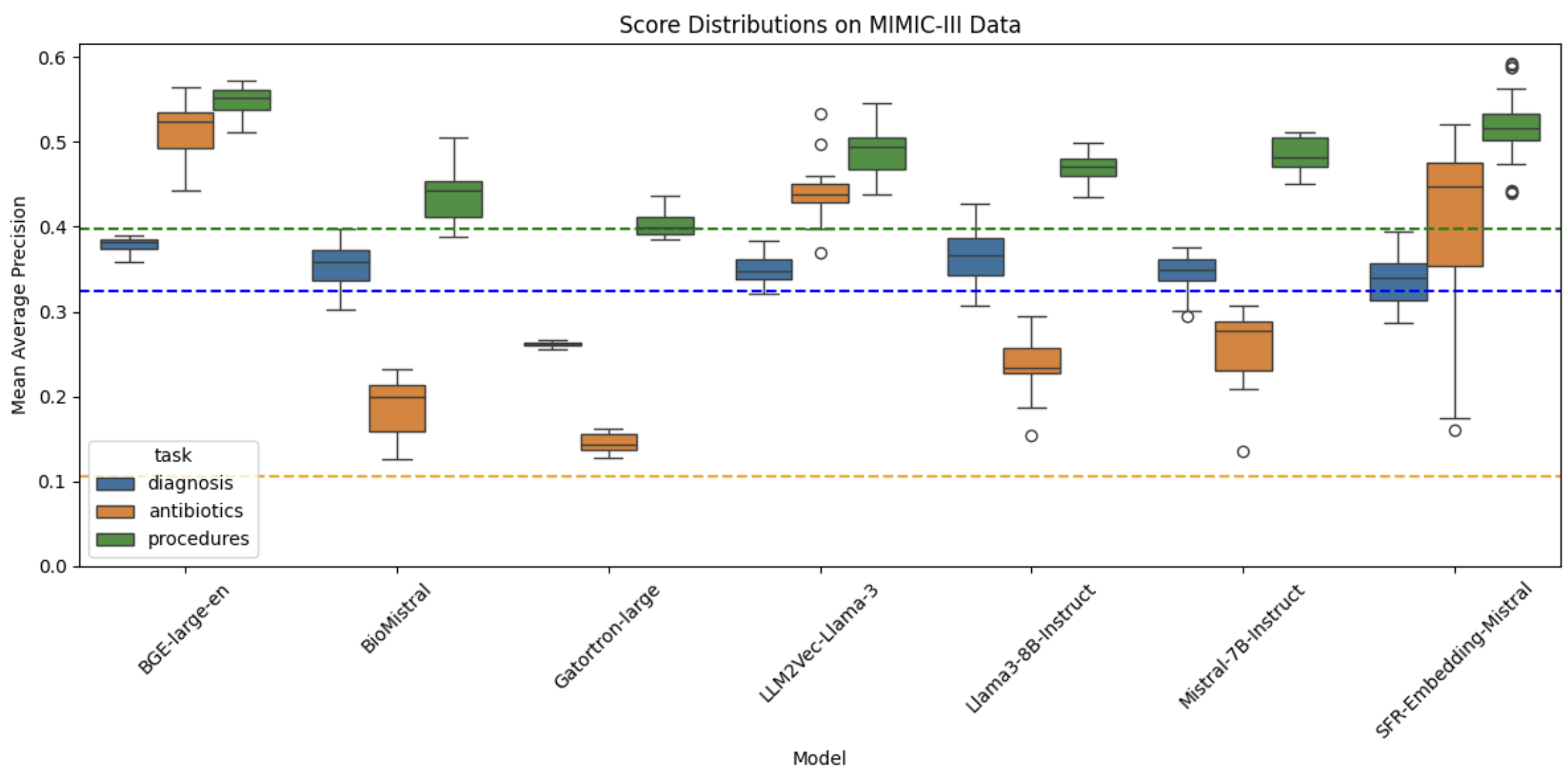}
  \caption{Boxes represent the interquartile range (IQR) of mean average precision scores for different query/query pooling samples for the different tasks and models on the MIMIC-III data, with the median marked. Whiskers extend to 1.5*IQR; outliers are shown as individual points. The dashed line is a baseline of random ordering of note chunks.}
  \label{fig:mimic_results}
\end{figure}

\begin{figure}[H]
  \centering
  \includegraphics[width=\textwidth]{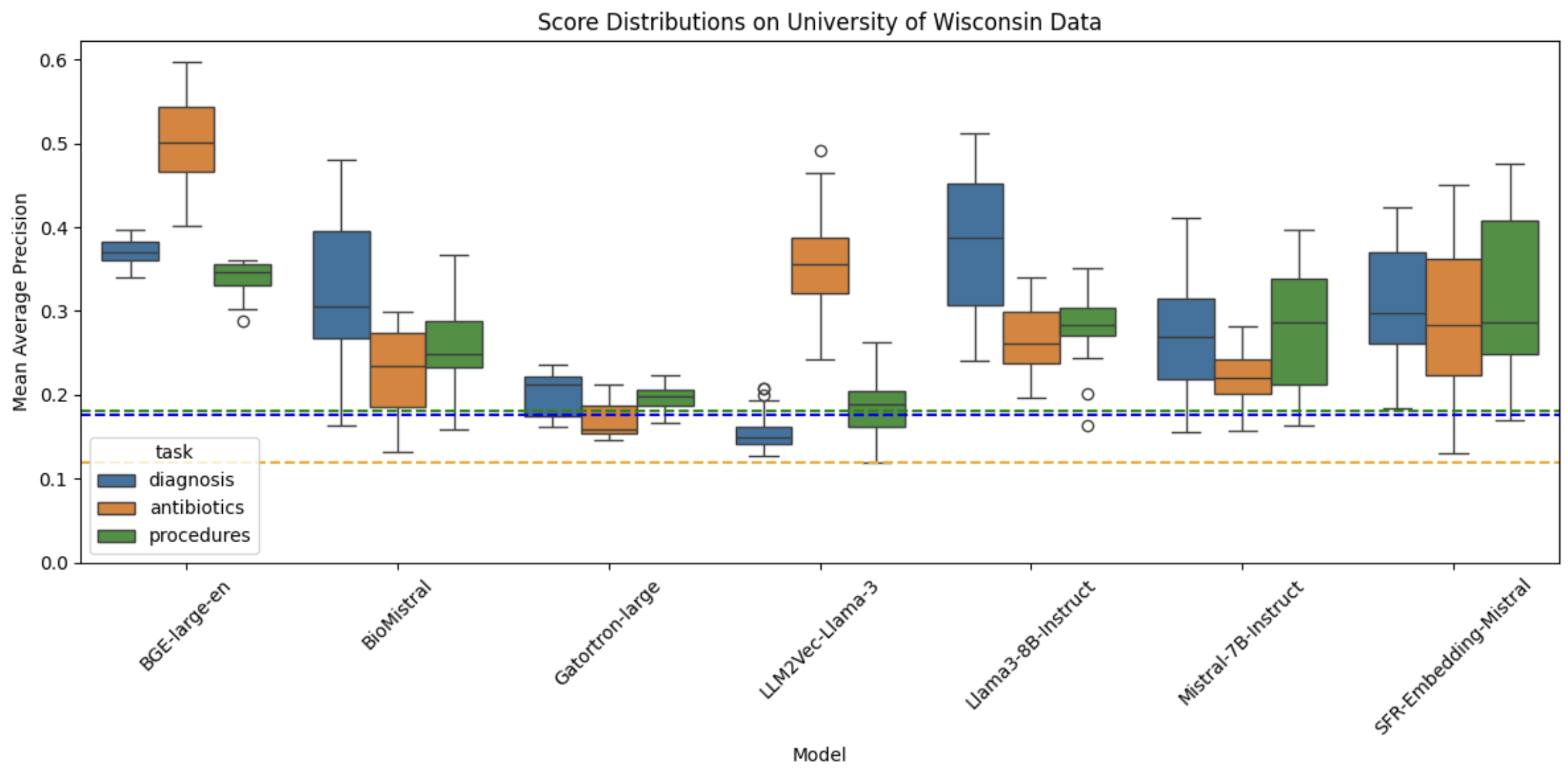}
  \caption{Boxes represent the interquartile range (IQR) of mean average precision scores for different query/query pooling samples for the different tasks and models on the UW data, with the median marked. Whiskers extend to 1.5*IQR; outliers are shown as individual points. The dashed line is a baseline of random ordering of note chunks.}
  \label{fig:private_results}
\end{figure}

We found that performance was very sensitive to the phrasing of the query, potentially even dropping it below baseline. For instance, using \textit{``primary diagnosis''} with Llama3-8B-Instruct and mean pooling, the average precision was 27.44. Simply changing the query to \textit{``patient's primary diagnosis''} drastically improved retrieval to 36.68. In Figures \ref{fig:mimic_results} and \ref{fig:private_results} we present box plots to illustrate the distribution of scores for each model and task. We observed that the UW dataset experiences more variability compared to MIMIC, regardless of model or task.

\section{Discussion}
\label{discussion}
In this study, we examined the impact of embedding methods for a RAG framework by examining various language models for embedding a corpus of clinical text and pooling methods for information retrieval on clinical tasks using both private and publicly available datasets. Our results demonstrated that BGE significantly outperformed all other models tested, despite scoring lower on the MTEB benchmark compared to SFR-Embedding-Mistral (54.29 vs. 59 on retrieval tasks) and LLM2Vec-Llama-3 (56.63), as well as being smaller than all other models tested. This discrepancy between benchmark performance and our evaluation underscores the importance of domain-specific assessments when deploying models in new contexts.

We also found substantial variability in the success of different queries, which highlights the need for tuning the queries themselves when conducting information retrieval and setting up RAG frameworks. Additionally, the more pronounced variability on the UW dataset compared to MIMIC-III warrants further exploration to understand the factors contributing to this difference.

Given the resources and time needed to perform over 3,000 experiments, we left several components of an RAG framework for future work. One important factor is the decision on how to break the data into chunks. Typical approaches include segmenting based on formatting (such as headers and paragraph breaks) or simply choosing a chunk size that fits within the embedding model's context limit. The length of these segments may significantly impact the performance of retrieval, either due to models' capability of representing larger amounts of text or their downstream effect on a generative model once retrieved.

\subsection{Limitations}
There are many other popular models for embeddings that we did not test, such as those in the GTE family \citep{li2023towards}, as well as other medical-domain models, such as Meditron \citep{chen2023meditron70b}. Due to legal and ethical restrictions on sharing the EHR data we use, we were unable to test on many of the popular closed-source models that are currently used, such as OpenAI's text-embedding family of models \citep{openai-embeddings} or Voyage AI's \citep{voyageaiEmbeddings}. Furthermore, for this reason, the UW data we evaluated on cannot be publicly released for the community to reproduce our results. 

\section{Conclusions}

In conclusion, our ablation study underscores the importance of carefully selecting and evaluating components when designing retrieval systems for the clinical domain. The choice of embedding model, pooling strategy, and query formulation can significantly impact retrieval performance, and further empirical studies like this one are crucial for making informed decisions that guide us toward more robust and effective retrieval systems. As the information in EHRs continues to grow exponentially, retrieval systems and vector databases that are scalable and reproducible in quality are becoming a viable solution to the information overload and note bloat problem \citep{patterson2024call}. Our initial work highlights the variants that can occur in the embedding quality and indexing for the later generative component of a RAG framework.

\section*{Funding acknowledgement}
This work was supported by the National Library of Medicine of the National Institutes of Health under award number R01LM012973. The content is solely the responsibility of the authors and does not necessarily represent the official views of the National Institutes of Health.

\section*{CRediT authorship contribution statement}
\textbf{Skatje Myers}:  Writing-original draft, Methodology, Data curation, Investigation, Conceptualization, Formal analysis, Software. \textbf{Timothy A. Miller}:  Writing–review \& editing, Methodology, Conceptualization, Funding acquisition, Supervision. \textbf{Yanjun Gao}:  Methodology, Writing–review \& editing. \textbf{Matthew Churpek}:  Methodology, Writing–review \& editing. \textbf{Anoop Mayampurath}:  Methodology, Writing–review \& editing. \textbf{Dmitriy Dligach}:  Conceptualization, Methodology, Writing–review \& editing, Supervision. \textbf{Majid Afshar}:  Supervision, Methodology, Formal analysis, Writing–review \& editing, Conceptualization, Funding acquisition.

\section*{Declaration of competing interest}

The authors declare that they have no known competing financial interests or personal relationships that could have appeared to influence the work reported in this paper.

\section*{Data availability}

The data underlying this article cannot be shared publicly due to privacy and data use agreements. The encounter IDs used for the MIMIC-III portion of the work will be shared on reasonable request to the corresponding author.

\bibliography{anthology,custom}

\begin{thebibliography}{10}

\bibitem{voyageaiEmbeddings}
Voyage AI.
\newblock {E}mbeddings --- docs.voyageai.com.
\newblock \url{https://blog.voyageai.com/2024/05/05/voyage-large-2-instruct-instruction-tuned-and-rank-1-on-mteb/}, 2024.
\newblock [Accessed 15-06-2024].

\bibitem{llama3modelcard}
AI@Meta.
\newblock Llama 3 model card.
\newblock 2024.

\bibitem{10.1007/978-3-031-53963-3_36}
Roman Aperdannier, Melanie Koeppel, Tamina Unger, Sigurd Schacht, and Sudarshan~Kamath Barkur.
\newblock Systematic evaluation of different approaches on embedding search.
\newblock In Kohei Arai, editor, {\em Advances in Information and Communication}, pages 526--536, Cham, 2024. Springer Nature Switzerland.

\bibitem{behnamghader2024llm2vec}
Parishad BehnamGhader, Vaibhav Adlakha, Marius Mosbach, Dzmitry Bahdanau, Nicolas Chapados, and Siva Reddy.
\newblock {LLM2Vec}: Large language models are secretly powerful text encoders, 2024.

\bibitem{canales2021assessing}
Lea Canales, Sebastian Menke, Stephanie Marchesseau, Ariel D’Agostino, Carlos del Rio-Bermudez, Miren Taberna, Jorge Tello, et~al.
\newblock Assessing the performance of clinical natural language processing systems: development of an evaluation methodology.
\newblock {\em JMIR Medical Informatics}, 9(7):e20492, 2021.

\bibitem{chen2023meditron70b}
Zeming Chen, Alejandro Hernández-Cano, Angelika Romanou, Antoine Bonnet, Kyle Matoba, Francesco Salvi, Matteo Pagliardini, Simin Fan, Andreas Köpf, Amirkeivan Mohtashami, Alexandre Sallinen, Alireza Sakhaeirad, Vinitra Swamy, Igor Krawczuk, Deniz Bayazit, Axel Marmet, Syrielle Montariol, Mary-Anne Hartley, Martin Jaggi, and Antoine Bosselut.
\newblock Meditron-70b: Scaling medical pretraining for large language models, 2023.

\bibitem{gao2023retrieval}
Yunfan Gao, Yun Xiong, Xinyu Gao, Kangxiang Jia, Jinliu Pan, Yuxi Bi, Yi~Dai, Jiawei Sun, and Haofen Wang.
\newblock Retrieval-augmented generation for large language models: A survey.
\newblock {\em arXiv preprint arXiv:2312.10997}, 2023.

\bibitem{jiang2023mistral}
Albert~Q Jiang, Alexandre Sablayrolles, Arthur Mensch, Chris Bamford, Devendra~Singh Chaplot, Diego de~las Casas, Florian Bressand, Gianna Lengyel, Guillaume Lample, Lucile Saulnier, et~al.
\newblock Mistral 7b.
\newblock {\em arXiv preprint arXiv:2310.06825}, 2023.

\bibitem{johnson2016mimic}
Alistair~EW Johnson, Tom~J Pollard, Lu~Shen, Li-wei~H Lehman, Mengling Feng, Mohammad Ghassemi, Benjamin Moody, Peter Szolovits, Leo Anthony~Celi, and Roger~G Mark.
\newblock {MIMIC-III}, a freely accessible critical care database.
\newblock {\em Scientific data}, 3(1):1--9, 2016.

\bibitem{labrak2024biomistral}
Yanis Labrak, Adrien Bazoge, Emmanuel Morin, Pierre-Antoine Gourraud, Mickael Rouvier, and Richard Dufour.
\newblock {BioMistral}: A collection of open-source pretrained large language models for medical domains, 2024.

\bibitem{lewis2020retrieval}
Patrick Lewis, Ethan Perez, Aleksandra Piktus, Fabio Petroni, Vladimir Karpukhin, Naman Goyal, Heinrich K{\"u}ttler, Mike Lewis, Wen-tau Yih, Tim Rockt{\"a}schel, et~al.
\newblock Retrieval-augmented generation for knowledge-intensive {NLP} tasks.
\newblock {\em Advances in Neural Information Processing Systems}, 33:9459--9474, 2020.

\bibitem{li2023towards}
Zehan Li, Xin Zhang, Yanzhao Zhang, Dingkun Long, Pengjun Xie, and Meishan Zhang.
\newblock Towards general text embeddings with multi-stage contrastive learning.
\newblock {\em arXiv preprint arXiv:2308.03281}, 2023.

\bibitem{liu2024lost}
Nelson~F Liu, Kevin Lin, John Hewitt, Ashwin Paranjape, Michele Bevilacqua, Fabio Petroni, and Percy Liang.
\newblock Lost in the middle: How language models use long contexts.
\newblock {\em Transactions of the Association for Computational Linguistics}, 12:157--173, 2024.

\bibitem{SFRAIResearch2024}
Rui Meng, Ye~Liu, Shafiq~Rayhan Joty, Caiming Xiong, Yingbo Zhou, and Semih Yavuz.
\newblock {SFR-Embedded-Mistral}.
\newblock Salesforce AI Research Blog, 2024.

\bibitem{muennighoff2022mteb}
Niklas Muennighoff, Nouamane Tazi, Lo{\"\i}c Magne, and Nils Reimers.
\newblock {MTEB}: Massive text embedding benchmark.
\newblock {\em arXiv preprint arXiv:2210.07316}, 2022.

\bibitem{patterson2024call}
Brian~W Patterson, Daniel~J Hekman, Frank~J Liao, Azita~G Hamedani, Manish~N Shah, and Majid Afshar.
\newblock Call me dr ishmael: trends in electronic health record notes available at emergency department visits and admissions.
\newblock {\em JAMIA open}, 7(2):ooae039, 2024.

\bibitem{10.1093/jamia/ocae029}
François Remy, Kris Demuynck, and Thomas Demeester.
\newblock {{BioLORD}-2023: semantic textual representations fusing large language models and clinical knowledge graph insights}.
\newblock {\em Journal of the American Medical Informatics Association}, page ocae029, 02 2024.

\bibitem{sainz-etal-2023-nlp}
Oscar Sainz, Jon Campos, Iker Garc{\'\i}a-Ferrero, Julen Etxaniz, Oier~Lopez de~Lacalle, and Eneko Agirre.
\newblock {NLP} evaluation in trouble: On the need to measure {LLM} data contamination for each benchmark.
\newblock In Houda Bouamor, Juan Pino, and Kalika Bali, editors, {\em Findings of the Association for Computational Linguistics: EMNLP 2023}, pages 10776--10787, Singapore, December 2023. Association for Computational Linguistics.

\bibitem{soldaini2016quickumls}
Luca Soldaini and Nazli Goharian.
\newblock Quick{UMLS}: a fast, unsupervised approach for medical concept extraction.
\newblock In {\em MedIR workshop, sigir}, pages 1--4, 2016.

\bibitem{wang2023improving}
Liang Wang, Nan Yang, Xiaolong Huang, Linjun Yang, Rangan Majumder, and Furu Wei.
\newblock Improving text embeddings with large language models.
\newblock {\em arXiv preprint arXiv:2401.00368}, 2023.

\bibitem{bge_embedding}
Shitao Xiao, Zheng Liu, Peitian Zhang, and Niklas Muennighoff.
\newblock C-pack: Packaged resources to advance general chinese embedding, 2023.

\bibitem{xiong2024benchmarking}
Guangzhi Xiong, Qiao Jin, Zhiyong Lu, and Aidong Zhang.
\newblock Benchmarking retrieval-augmented generation for medicine.
\newblock {\em arXiv preprint arXiv:2402.13178}, 2024.

\bibitem{yang2022large}
Xi~Yang, Aokun Chen, Nima PourNejatian, Hoo~Chang Shin, Kaleb~E Smith, Christopher Parisien, Colin Compas, Cheryl Martin, Anthony~B Costa, Mona~G Flores, et~al.
\newblock A large language model for electronic health records.
\newblock {\em NPJ digital medicine}, 5(1):194, 2022.

\bibitem{openai-embeddings}
Juntang Zhuang, Paul Baltescu, Joy Jiao, Arvind Neelakantan, Andrew Braunstein, Jeff Harris, Logan Kilpatrick, Leher Pathak, Enoch Cheung, Ted Sanders, Yutian Liu, Anushree Agrawal, Andrew Peng, Ian Kivlichan, Mehmet Yatbaz, Madelaine Boyd, Luisa Anna-Brakman, Leoni Florencia~Aleman, Henry Head, Molly Lin, Meghan Shah, Chelsea Carlson, Sam Toizer, Ryan Greene, Alison Harmon, Denny Jin, Karolis Kosas, Marie Inuzuka, Peter Bakkum, Barret Zoph, Luke Metz, Jiayi Weng, Randall Lin, Yash Patil, Mianna Chen, Andrew Kondrich, Brydon Eastman, Liam Fedus, John Schulman, Vlad Fomenko, Andrej Karpathy, Aidan Clark, and Owen Campbell-Moore.
\newblock {N}ew embedding models and {A}{P}{I} updates.
\newblock \url{https://openai.com/index/new-embedding-models-and-api-updates/}, 2024.
\newblock [Accessed 15-06-2024].

\end{thebibliography}
\bibliographystyle{plain}


\section{Appendix}
\label{sec:appendix}

We present the lists of queries used for the ``diagnosis'' (Table \ref{tab:diag_query}), ``procedures'' (Table \ref{tab:proc_query}), and ``antibiotics'' (Table \ref{tab:anti_query}) tasks. The queries that begin with ``Instruct:[...]'', ``Given [...]'', and ``Represent [...]'' were only used for SFR-Embedding-Mistral, LLM2Vec, and BGE (respectively). These models were trained to use these formats, although they are not strictly necessary.

\begin{table}[H]
\begin{tabular}{|l|}
\hline
The patient's primary diagnosis is                                                                                                                                          \\ \hline
patient's primary diagnosis                                                                                                                                                 \\ \hline
What is the patient's primary diagnosis?                                                                                                                                    \\ \hline
The patient has been diagnosed with                                                                                                                                         \\ \hline
primary diagnosis                                                                                                                                                           \\ \hline
\begin{tabular}[c]{@{}l@{}}Represent this sentence for searching relevant passages: \\ patient's primary diagnosis\end{tabular}                                             \\ \hline
\begin{tabular}[c]{@{}l@{}}Represent this sentence for searching relevant passages: \\ What is the patient's primary diagnosis?\end{tabular}                                \\ \hline
\begin{tabular}[c]{@{}l@{}}Given a search query, retrieve relevant passages that answer the query: \\ patient's primary diagnosis\end{tabular}                              \\ \hline
\begin{tabular}[c]{@{}l@{}}Given a search query, retrieve relevant passages that answer the query: \\ What is the patient's primary diagnosis?\end{tabular}                 \\ \hline
\begin{tabular}[c]{@{}l@{}}Instruct: Given a search query, retrieve relevant passages that answer the query.\\ Query: patient's primary diagnosis\end{tabular}              \\ \hline
\begin{tabular}[c]{@{}l@{}}Instruct: Given a search query, retrieve relevant passages that answer the query.\\ Query: What is the patient's primary diagnosis?\end{tabular} \\ \hline
\end{tabular}
\caption{Queries used for retrieving the primary diagnosis.}
\label{tab:diag_query}
\end{table}

\begin{table}[H]
\begin{tabular}{|l|}
\hline
The patient received the following surgical procedures:                                                                                                                                                            \\ \hline
surgical procedures                                                                                                                                                                                                \\ \hline
What invasive and surgical procedures did the patient receive?                                                                                                                                                     \\ \hline
surgical procedures and operations                                                                                                                                                                                 \\ \hline
\begin{tabular}[c]{@{}l@{}}Represent this sentence for searching relevant passages:\\ What invasive and surgical procedures did the patient receive?\end{tabular}                                                  \\ \hline
\begin{tabular}[c]{@{}l@{}}Represent this sentence for searching relevant passages:\\ invasive and surgical procedures?\end{tabular}                                                                               \\ \hline
\begin{tabular}[c]{@{}l@{}}Given a search query, retrieve relevant passages that answer the query: \\ What invasive and surgical procedures did the patient receive?\end{tabular}                                  \\ \hline
\begin{tabular}[c]{@{}l@{}}Given a search query, retrieve relevant passages that answer the query: \\ invasive and surgical procedures?\end{tabular}                                                               \\ \hline
\begin{tabular}[c]{@{}l@{}}Instruct: Given a search query, retrieve relevant passages that answer the query.\textbackslash{}n\\ Query: What invasive and surgical procedures did the patient receive?\end{tabular} \\ \hline
\begin{tabular}[c]{@{}l@{}}Instruct: Given a search query, retrieve relevant passages that answer the query.\textbackslash{}n\\ Query: invasive and surgical procedures?\end{tabular}                              \\ \hline
\end{tabular}
\caption{Queries used for retrieving the invasive and surgical procedures.}
\label{tab:proc_query}
\end{table}

\begin{table}[H]
\begin{tabular}{|l|}
\hline
antibiotics                                                                                                                                                                                \\ \hline
antibiotic medications                                                                                                                                                                     \\ \hline
What antibiotics are the patient taking?                                                                                                                                                   \\ \hline
The antibiotics the patient is taking are                                                                                                                                                  \\ \hline
\begin{tabular}[c]{@{}l@{}}Represent this sentence for searching relevant passages:\\ What antibiotics has the patent taken?\end{tabular}                                                  \\ \hline
\begin{tabular}[c]{@{}l@{}}Represent this sentence for searching relevant passages:\\ antibiotics\end{tabular}                                                                             \\ \hline
\begin{tabular}[c]{@{}l@{}}Given a search query, retrieve relevant passages that answer the query: \\ What antibiotics has the patent taken?\end{tabular}                                  \\ \hline
\begin{tabular}[c]{@{}l@{}}Given a search query, retrieve relevant passages that answer the query:\\ antibiotics\end{tabular}                                                              \\ \hline
\begin{tabular}[c]{@{}l@{}}Instruct: Given a search query, retrieve relevant passages that answer the query.\textbackslash{}n\\ Query: What antibiotics has the patent taken?\end{tabular} \\ \hline
\begin{tabular}[c]{@{}l@{}}Instruct: Given a search query, retrieve relevant passages that answer the query.\textbackslash{}n\\ Query: antibiotics\end{tabular}                            \\ \hline
\end{tabular}
\caption{Queries used for retrieving mentions of antibiotics.}
\label{tab:anti_query}
\end{table}

\end{document}